\documentclass{article}
\usepackage{ijcai19}
\usepackage{colortbl}
\usepackage{times}
\usepackage{soul}
\usepackage{url}
\usepackage[utf8]{inputenc}
\usepackage[small]{caption}
\usepackage{graphicx}
\usepackage{booktabs}
\usepackage{mathrsfs}
\usepackage{comment}
\usepackage{algorithm}
\usepackage{algpseudocode}
\usepackage{multirow}
\usepackage{amsmath}
\usepackage{color}
\newcommand{\xw}[1]{\textcolor{blue}{#1}}

\definecolor{mygray}{gray}{.9}
\newcommand{\bg}[0]{\cellcolor{mygray}}

\title{ Amalgamating  Filtered Knowledge: \\Learning
Task-customized Student from Multi-task Teachers}

\author{
Jingwen Ye$^1$
\and
Xinchao Wang$^2$\and
Yixin Ji$^1$\and
Kairi Ou$^3$\And
Mingli Song$^1$
\affiliations
$^1$College of Computer Science and Technology, Zhejiang University, Hangzhou, China\\
$^2$Department of Computer Science, Stevens Institute of Technology, New Jersey, United States\\
$^3$Alibaba Group, Hangzhou, China\\
\emails
\{yejingwen, jiyixin, brooksong\}@zju.edu.cn,
xinchao.w@gmail.com,
suzhe.okr@taobao.com
}

\begin{document}

\maketitle

\begin{abstract}
Many well-trained Convolutional Neural Network~(CNN) models have now been released online by developers
for the sake of effortless reproducing.
In this paper, we treat such pre-trained networks as teachers, and explore how to learn a target student network for customized tasks, using multiple teachers that handle different tasks.
We assume no human-labelled annotations are available, and each teacher model can be either single- or multi-task network, where the former is a degenerated case of the latter.
The student model, depending on the customized tasks, learns the related knowledge filtered from the multiple teachers, and eventually masters
the complete or a subset of expertise from all teachers.
To this end, we adopt a layer-wise training strategy, which entangles the student's network block to be learned with the corresponding teachers.
As demonstrated on several benchmarks, the learned student network achieves
very promising results, even outperforming the teachers on the customized tasks.

\end{abstract}

\section{Introduction}
{In recent years, deep learning has achieved state-of-the-art performances in most if not all of the artificial intelligent applications.} The success of deep learning, however, heavily relies on a massive number of human annotations, sometimes even up to the scales of tens of millions such as those of ImageNet. In many cases, the training data or annotations are confidential and therefore not available to the public. To alleviate the re-training and reproducing effort, many developers have now released their trained models online, so that users may use them directly or finetune them on a new dataset.

\begin{figure}[t]
\centering
\includegraphics[scale = 0.55]{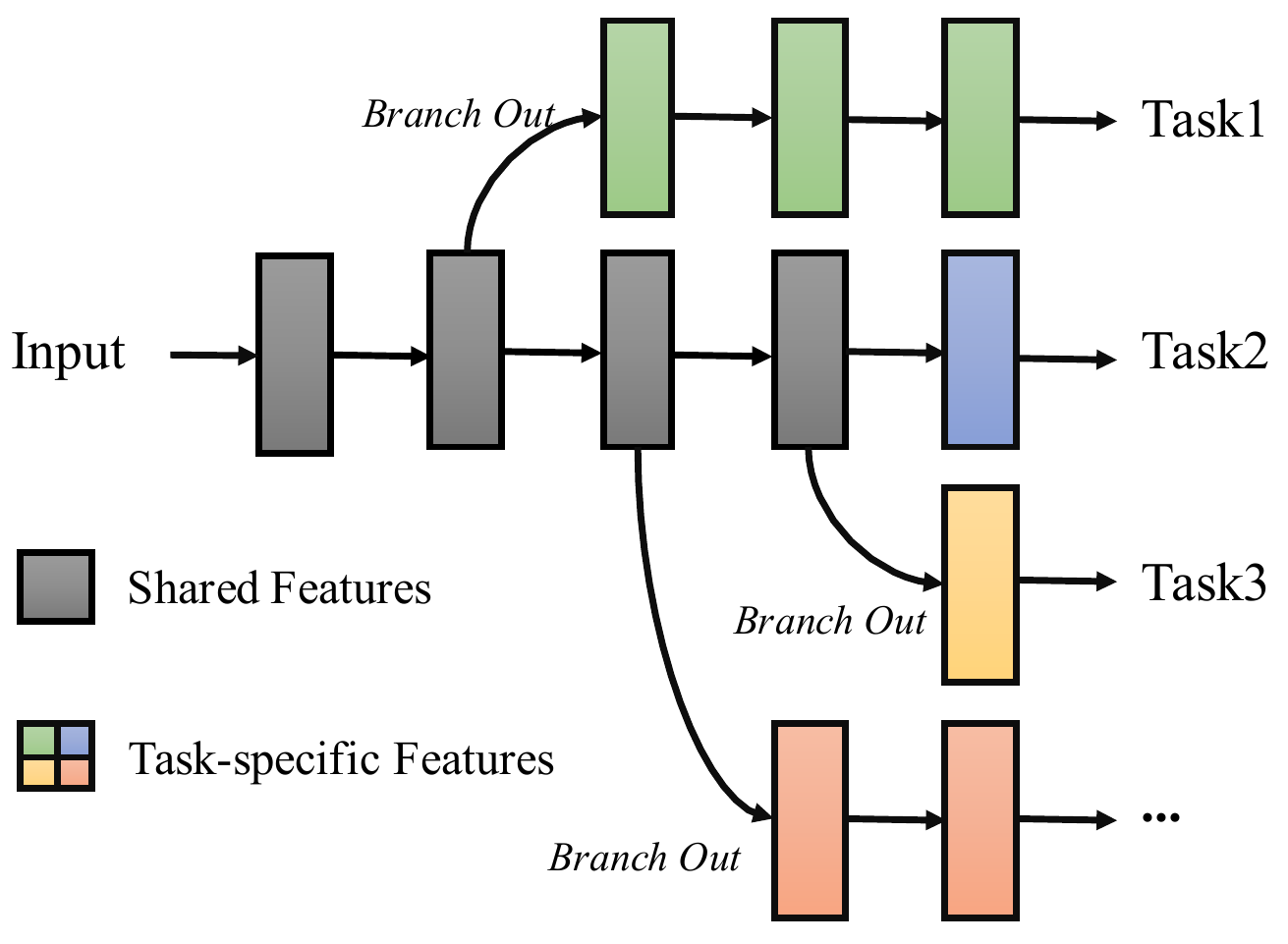}
\caption{ The TargetNet learns to solve customized multiple tasks. The features learned in TargetNet are divided into two parts, the shared features shown in black and the task-specific features in colors. Branching out takes place at different layers for distinct tasks.}
\label{fig:task}
\end{figure}

In this paper, we study how to take advantage of such pre-trained networks to produce a new and versatile network, without having access to any human annotation. Our problem setup is as follows. We assume that we are given a pool of pre-trained networks, like those downloaded from the model zoo, as teacher models. Each such teacher may be either a single- or multi-task model, where the former  can be treated as a degenerated case of the latter. We aim to train a customized multi-task student model, termed TargetNet, which masters an arbitrary set of the teachers' expertise, using only the provided teacher models and without human annotations. In other words, the student amalgamates the multi-domain knowledge, relevant to its pre-designated customized task, from the multiple teachers that specialize in different tasks.

The proposed task therefore finds its important application in CNN reusing, as it requires no human annotation yet aggregates the knowledge and consequently handles the tasks of multiple teachers of distinct domains, each of which, again, may be single- or multi-task.
It is more generalized than  Knowledge Distilling~(KD) that learns a student working on the same task as a single teacher does, and {more flexible than multi-task learning since we can customize the tasks in need.}
As the initial attempt towards this aspiring goal,  for this work we focus on classification and utilize the state-of-the-art ResNet~\cite{he2016identity} and DenseNet~\cite{huang2017densely}
backbone for the student and the teachers. Nevertheless, the proposed approach is not restricted to these two backbones and others can be applied as well.

An example of the learned multi-task student network is depicted in Fig.~\ref{fig:task}, where the first several layers shown in black correspond to shared features and each particular task branches out at its own individual location.
Specifically, the TargetNet is learned in a layer-wise manner, where we go through each layer and amalgamate the filtered knowledge from those teachers relevant to the customized tasks. This is achieved via entangling the layer of the student with the teacher network, in other words, by replacing the corresponding layer in the teacher with the one to be learned in the student, followed by computing the loss and eventually putting the layer back in the student. The branch-out location for each individual task, on the other hand, is  obtained by searching for the layer that yields the overall minimum training loss.

With the proposed training strategy, we are able to derive a TargetNet that, not only masters a combination of various teachers' skills but also outperforms the teachers in their own domains. Furthermore, the size of the TargetNet is much smaller than the ensemble of the relevant teachers.

Our contribution is therefore an effective approach to training a student model termed TargetNet, without human annotations, that amalgamates knowledge from a pool of multi- or single-task teachers working on different tasks. From these teachers we filter the relevant knowledge to train the TargetNet by entangling them all.
The TargetNet after training, at a reasonably compact size, handles multiple customized tasks and leads to results superior to the corresponding teachers.

\section{Related Work}
In this section, we first briefly review the recent approaches in multi-task learning, and then look at a related task, knowledge distillation, which focuses on training a student model that handles the same task as the teachers do.


\textbf{Multi-task Learning.}
Multi-task learning has been widely studied in
computer vision,
natural language processing,
and machine learning
communities~\cite{GongICML16}.
To describe the hierarchical relations among tasks and
to learn model parameters under the regularization frameworks, several lines of work
have been proposed.

One line is to utilize the given tree structure. For example, \cite{Han2014Encoding} proposes a probabilistic tree sparsity model that utilizes the tree structure to obtain the sparse solution instead of the group structure. More recently, ML-Forest~\cite{wu2016ml-forest:} is proposed to learn an ensemble of hierarchical multi-label classifier trees to reveal the intrinsic label dependencies.

Another line has been focused on fusing MTL with CNN to learn the shared features and the task-specific models.
For example,~\cite{zhang2014facial} proposes a deep CNN for joint face detection, pose estimation, and landmark localization. Misra et. al.~\cite{misra2016cross-stitch} propose a cross-stitch network for MTL to learn an optimal combination of shared and task-specific representations. In~\cite{zhang2014improving}, a task-constrained deep network is developed for landmark detection with facial attribute classifications as the side tasks.~\cite{Zhao2018A} proposes a multi-task learning system to jointly train the task of image captioning and two other related auxiliary tasks which help to enhance the CNN encoder and the RNN decoder in the image captioning model.


Considering a large number of possible label sets, most multi-task learning methods require sufficient labeled training samples. Multi-Label Co-Training~\cite{yuying2018multi} introduces a semi-supervised method which leverages information concerning the co-occurrence of pairwise labels.

Unlike traditional multi-task learning, our customized multi-task learns the tasks' hierarchy in CNN without the new demand for dataset with groundtruth annotations.

\textbf{Knowledge Distillation.}
First proposed in~\cite{hinton2015distilling}, knowledge
distillation aims at training a student model of a compact size
by learning from a larger teacher model or a set of teachers handling
the same task, and thus finds its important application in deep model
compression~\cite{WangCVPR17}.
More recently, the work of~\cite{gao2017knowledge} introduces a multi-teacher and single-student knowledge concentration approach. The work of~\cite{shen2019amalgamating},
on the other hand, trains a student classifier by learning from multiple teachers working on different classes. Each teacher here, however, is restricted to be single-task.

There are also attempts focusing on tasks other than classification.
For example,~\cite{chen2017learning} handles object detection and
learns a student model with improved accuracy;~\cite{huang2018knowledge}
specializes in speech applications by conducting knowledge distillation
on sequence data.

The proposed work here, unlike any prior ones we are aware of,
trains a student model by learning from a pool of teachers,
each of which may be either single- or multi-task.
The student model, under the specification of users,
and amalgamates only the \emph{filtered} knowledge.

\section{Method}
In this section, we provide the technical details of the proposed approach.
\begin{figure*}[t]
\centering
\includegraphics[scale = 0.62]{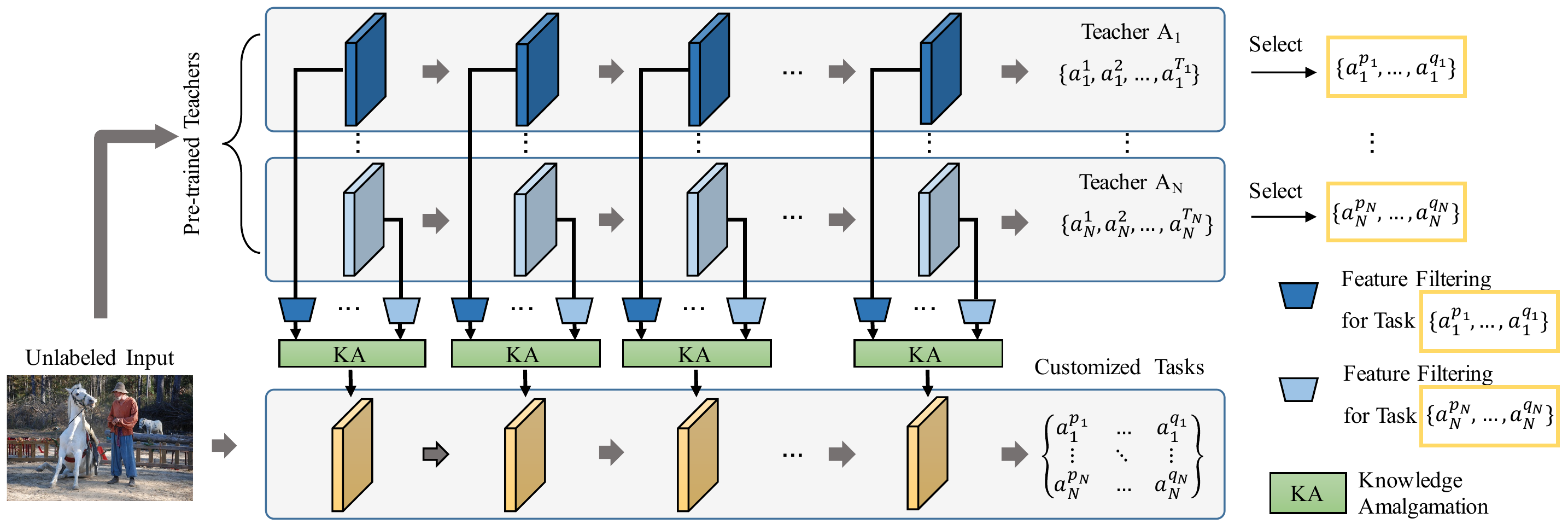}
\caption{Task-customized knowledge amalgamation in the target network. Knowledge amalgamation happens in each block of the networks (teacher networks and target network). $N$ teachers' knowledge is utilized for the guidance of TargetNet.}
\label{fig:ka}
\end{figure*}

\subsection{Problem Definition}
Our goal is to train a student model, termed as TargetNet, of which the tasks are customized by the users.
As depicted in Fig.~\ref{fig:ka},
we learn TargetNet using
pre-trained teacher models and unlabelled images.
Since each teacher may handle multiple tasks, of which only a subset is relevant to TargetNet,
we would have to first \emph{filter} the knowledge from a teacher and then \emph{amalgamate}
such filtered knowledge from all the teachers so as to produce TargetNet.

Let us use $I$ to denote an input image,
and use
$\mathcal{A}=\{\mathcal{A}_1,\mathcal{A}_2,...,\mathcal{A}_N\}$
to denote a set of teachers, each of which may be multi-task.
Let $T_n$  denote the number of tasks handled by a teacher $\mathcal{A}_n$,
and thus we have
$A_n=\{a_n^1,a_n^2,...,a_n^{T_n}\}$, forming the teacher tasks' union as $A=\{A_1,...,A_N\}$.
To make a customized task union for target, tasks $\{a_n^{p_n},...,a_n^{q_n}\}$ are selected in teacher $n$. Then for TargetNet, the customized tasks can be defined as:
\begin{equation}
\begin{split}
&C_n=\{a_n^{p_n},...,a_n^{q_n}\}\\
&C=\{C_1,C_2,...,C_N\},
\end{split}
\end{equation}
where $C_n\subseteq A_n$, $C\subseteq A$, which means the customized tasks can either be the whole union or a subset of the teachers' task ensemble.

In our approach, we take a \emph{block} to be the atomic unit of a network.
Specifically, we use $F_n^k$ to denote the
feature maps in the $k$-th block of the $n$-th pre-trained teacher, which are
to be filtered first before being further amalgamated.

\subsection{ Customized Multi-task TargetNet}

In this section, we describe the proposed approach to learning a compact student network, TargetNet. The training is achieved via a strategy, which learns the parameters of the student entangled with those of the teachers.
In essence, our approach is a block-wise learning method, as depicted in Fig.~\ref{fig:ka}, that learns the parameters of the student network through
transforming the student knowledge into each of the teachers' expertise domains for computing the loss and thus updating the parameters.

The regions with high activations from a neuron may share some task related similarities, even though these similarities may not be intuitive for human interpretation. In order to capture these similarities, there should be also neurons mimic these activation patterns in student networks.

Here, to clarify the proposed method, we describe the details in the multi-label task, which is a specific field of multi-task problem. Despite the TargetNet eventually has multiple output streams for each task, it is initialized with the same architecture as those of the teachers except for the last few fully connected layers(fc), where we set the number of fc the same as that of tasks. We then train the TargetNet and finally branch out the streams for each task, after which the initial blocks following the branch-out are removed.

\subsection{Pretrained Teacher Networks}
In the set of $N$ pretrained teacher networks with the same architecture $\mathcal{A}$, each teacher $\mathcal{A}_n$ specializes in $T_n$ different tasks, where $T_n\ge 1$ which means the teachers can be either in single or multiple tasks' architecture.

For teacher network $\mathcal{A}_n$ with $T_n$ tasks, when $T_n=1$ with task $A_n=\{a_n^1\}$, then the parameters for network $\mathcal{A}_n$ can be easily updated with the loss $\ell(a_n^1,a_n^{1*})$. In the situation that $T_n>1$, and teacher $\mathcal{A}_n$ is the expert for task $A_n=\{a_n^1,a_n^2,...,a_n^{T_n}\}$. Then the loss function can be written as:
\begin{equation}
\label{eq:teacherloss}
l_n=\frac{1}{T_n}\sum_i \ell(a_n^i,a_n^{i*})
\end{equation}
where $\ell$ is the cross entropy loss for each task in network $\mathcal{A}_n$, $a_n^i$ and $a_n^{i*}$ are the groudtruths and predict labels for $i$-th task respectively.

We use this loss function to pre-train the network to get parameters $\theta$. Given an input image $I$, the outputs of $T_n$ tasks can be defined as $\mathcal{O}_n^{(T_n)}=\mathcal{A}_n(I,\theta_n)$. For each teacher network, we define the \emph{block} to be the minimum unit, then the network can be denoted as a stack of $B$ blocks $\mathcal{A}_n=\{\mathcal{A}_n^1,\mathcal{A}_n^2,...,\mathcal{A}_n^B\}$ and the last few fully connected layers ${\mathcal{F}}_n^{(T_n)}$. The feature maps output from each block can be defined as:
\begin{equation}
\begin{split}
&F_n^k= \mathcal{A}_n^k([F_n^1,...,F_n^{k-1}],\theta_n^k) \quad 1<k\le B\\
&\mathcal{O}^{(T_n)} ={\mathcal{F}}_n^{(T_n)}(F_n^1,...,F_n^{B});
\end{split}
\end{equation}
where $F^1 = \mathcal{A}^1(I,\theta^1)$, which takes the original image as input.

In this paper, $F^k$ contains the knowledge we need for guiding the target network. When the tasks in the teacher are all selected for the target, the intermediate features $\{F^1,F^2,...,F^B\}$ contain all the knowledge to be amalgamated for TargetNet. When only a part of the tasks are chosen, $F^{(B)}$ can't be directly used for guidance for the reason that the knowledge of the un-selected tasks is entangled with no difference and will pollute the customized tasks.

\subsection{Filtered-Knowledge Amalgamation}

For training the customized multi-task TargetNet $\mathcal{T}$ for the selected task set ${C}$, we transfer the problem into learning the features $F_u^k$ at the $k$-th block of TargetNet. We feed the unlabelled samples into both teachers and the target, in which way the features of the teachers and the initial ones of the Target are both obtained. To amalgamate the useful part of them, we introduce a two-step filtering strategy to acquire the exactly needed knowledge for the chosen customized tasks, one is the teacher-level filtering $f$, the other is the task-level one $g$.

For teacher-level filtering, considering the learned knowledge $F_u$ in TargetNet works as a container having the knowledge of all the teachers, we need to decompose it into $N$ streams as $F_{un}$ for each teacher. Then the process can be defined as
\begin{equation}
    F_{un}^k=f_n^k(F_u^k);
\end{equation}
which is applied in every block $k$, with $1\le n\le N$ and $1\le k \le B$. This filtering function is learnable and realized by a light channel coding module consisting of a global pooling layer and two fully connected layers.

To deal with the task-customized demand, a task-level filtering is constructed. Let $F_n^k$ denotes the obtained features at the $k$-th block of the $n$-th teacher network. Obviously, this intermediate features are learned in pre-trained networks for all the tasks $\{a_n^1,a_n^2,...,a_n^{T_n}\}$. So for the customized task set $C_n$, we use $g$ to filter the features specifying for the selected tasks, then $g(F_n^k)$ outputs the customized knowledge, where $g(F_n^k)=F_n^k$ when $C_n=A_n$,
And the loss function to update block $k$ in TargetNet can be defined as:
\begin{equation}
\label{eq:loss}
\mathcal{L}_u^k=\sum_n\lambda_nDist\{F_{un}^k,g(F_n^k)\},
\end{equation}
where $\lambda_n$ is a set of weights that balances each network, ${Dist}(\cdot)$ is the distance function measuring the difference between the features.

With the loss function Eq.~(\ref{eq:loss}), parameter updating takes place within block $k$, in $\mathcal{T}^k$ and the connecting teacher-level filters $\{f_n^k\}_{n=1}^{N}$, and blocks  $\{\mathcal{T}^1,...,\mathcal{T}^{k-1}\}$ keep unchanged.

However, since the knowledge of the teachers and the target are in the different feature space, the distance function is hard to determine for directly aligning the features between teachers and student. Also, for training each block in TargetNet, we have to adjust the weights $\lambda_n$ and the termination conditions, as the magnitudes of the feature maps and convergence rates vary block by block. Moreover, without groundtruth annotations, the filtering function $g(\cdot)$ in Eq.~(\ref{eq:loss}) seems impossible to define and train.

To this end, we learn the features of the student
by entangling them with those of the teachers.
In block $n$, we compute $F_{un}^k$ by sending
$F_u^k$, denoting the amalgamated features, to
$f_n^k$, denoting the filter at the corresponding level of the teacher.
Next, we substitute
the features $F_n^k$ at the $k$-th block of $n$-th teacher
with $F_{un}^k$, and derive the predictions
$\{\mathcal{O}_{un}^{i}\}_{i=p_n}^{q_n}$ from the $n$-th teacher with $F_{un}^k$ as its features. We write,

\begin{equation}
\label{output}
\begin{split}
&\mathcal{O}_{un}^{i}=\mathcal{F}_n^{i}([F_n^1,...,F_n^{k-1},F_{un}^k,F_n^{k+1*},...,F_n^{B*}])\\
&F_n^{l*}=\mathcal{A}_n^{l}([F_n^1,...,F_n^{k-1},F_{un}^k,F_n^{k+1*},...,F_n^{l*}])\\
&F_n^{k+1*}=\mathcal{A}_n^{k+1}([F_n^1,...,F_n^{k-1},F_{un}^k]),
\end{split}
\end{equation}
where $k<l\le B$, $p_n\le i\le q_n$. In the final output, we only select the needed predictions for customized tasks $C_n$: $\{\mathcal{O}_{un}^{p_n},...,\mathcal{O}_{un}^{q_n}\}\subseteq\mathcal{O}_{un}^{(T_n)}$.

Then the losses with corresponding tasks are calculated and the loss $\mathcal{L}_u$ in Eq.~(\ref{eq:floss}) can be rewritten as:
\begin{equation}
\label{eq:floss}
    \mathcal{L}_u=\frac{1}{|C|}\sum_{n=1}^N\sum_{i=p_n}^{q_n}\ell(\mathcal{O}_n^i,\mathcal{O}_{un}^i),
\end{equation}
where $\ell$ in the same as the loss function in Eq.~(\ref{eq:teacherloss}). $|C|$ is the total number of tasks in the customized task set $C$, which is:
\begin{equation}
    |C|=\sum_n|C_n|=\sum_n(q_n-p_n+1).
\end{equation}



\subsection{Branch Out}
\label{sec:branchout}
As we are handling multiple tasks here,
it is non-trivial to decide where to branch out the TargetNet into separate task-specific streams to achieve the optimal performances on all tasks simultaneously.
Here, we allow different tasks to branch out at different spots.
Specifically, once the $B$ blocks of TargetNet have been trained using Eq.~(\ref{eq:floss}),
we obtain the  trained blocks $\{\mathcal{T}^1(\theta_u^1),\mathcal{T}^2(\theta_u^2),...,\mathcal{T}^B(\theta_u^B)\}$ for TargetNet $\mathcal{T}$  and the final loss values for each block and  each task $\mathcal{L}_n^i$. We then take the branch-out blocks to be:
\begin{equation}
\label{eq:branchout}
\begin{split}
S_n^i=\arg\min_n{\mathcal{L}_n^i}.
\end{split}
\end{equation}
Then for each task in the customized task set $C$, we can get a branch-out block spot.

\begin{algorithm}[hb]
  \caption{Training TargetNet with $n$-th teacher}
  \label{alg::whole}
  \begin{algorithmic}[1]
    \Require
      $I$: original input;
      $\mathcal{A}_n(\theta_n)$: $n$-th pre-trained teacher specifies in tasks $\{a_n^i\}_{i=1}^{T_n}$;
      $B$: the total number of blocks;
      $\{a_n^i\}_{i=p_n}^{q_n} $: customized tasks.
    \Ensure
      $\{\mathcal{T}_i\}_{i=p_n}^{q_n}$: task-specific TargetNet.
    \State Initial TargetNet $\mathcal{T}(\theta_0)$ and task-level filter $f_n$;
    \State Feed $I$ into $\mathcal{A}_n$ and $\mathcal{T}$, $\mathcal{O}_n^{(T_n)}=\mathcal{A}_n(I)$
    \State Initial $F_u^0=F_n^0=I$
    \For{block $k=1:B$}
      \State Get $F_u^k=\mathcal{T}^k([F_u^1,...,F_u^{k-1}],\theta_0^k)$;
      \State Transform $F_u^k$ to $F_{un}^k=f_n^k(F_u^k)$;
      \State Compute $ \mathcal{O}_{un}^{(T_n)}$ using Eq.~(\ref{output});
      \State Compute the loss of $\{\mathcal{O}_{un}^i\}_{i=p_n}^{q_n}$ and $\{\mathcal{O}_{n}^i\}_{i=p_n}^{q_n}$ ;
      \State Update $\mathcal{T}^k(\theta_u^k)$ and $f_n^k$;
      \State Get the convergence loss value $\mathcal{L}_n^i$.
    \EndFor
    \State Find branch-out point $S_n^i=\arg\min_n{\mathcal{L}_n^i}$;
    \State Get $\mathcal{T}_i$ by connecting $\{\mathcal{T}^1(\theta_u^1),...,\mathcal{T}^{S_n^i}(\theta_u^{S_n^i})\}$ with $\mathcal{C}_n^{S_n^i}$,  $\{\mathcal{A}_n^{S_n^i+1}(\theta_n^{S_n^i+1}),...,\mathcal{A}_n^B(\theta_n^B)\}$ and $\mathcal{F}_n^i$.
    \State Fine-tune $\{\mathcal{T}_i\}_{i=p_n}^{q_n}$ and return it.
  \end{algorithmic}
\end{algorithm}

We take out the initial blocks following the last branch-out block, once the branch-out spot is determined, meaning that all the initial blocks after $\max(S_n^i)$ are removed. Then for each task $i$ selected from teacher $\mathcal{A}_n$, we choose to replicate the corresponding following blocks in the teacher to that branch of task. That is, we regroup the branched task-specific TargetNet for task $i$ as:
\begin{equation}
\begin{split}
\mathcal{T}_i=[\{\mathcal{T}^1(\theta_u^1),\mathcal{T}^2(\theta_u^2),...,\mathcal{T}^{S_n^i}\},\mathcal{C}_n^{S_n^i},\\
\{\mathcal{A}_n^{S_n^i+1}(\theta_n^{S_n^i+1}),...,\mathcal{A}_n^B(\theta_n^B)\},\mathcal{F}_n^i].
\end{split}
\end{equation}
where we also keep the corresponding teacher-wise filter module for connecting the TargetNet and the corresponding teacher.

Each task-specific $\mathcal{T}_i$ shares the first few blocks of the TargetNet, and branches out at different blocks, which is the way to make them hierarchical.
As a whole, the process of training TargetNet from one teacher is concluded in Alg.~\ref{alg::whole}.

\section{Experiments and Results}
Here we provide our experimental settings and results. We test the performance on the multi-label problem.

\subsection{Experimental Settings}
\textbf{Datasets.} We evaluate the proposed method on the PASCAL Visual Object Classes (VOC) dataset~\cite{pascal-voc-2007}.
In this paper, both PASCAL VOC 2007 and VOC 2012 are employed for experiments. These two datasets contain 9,963 and 22,531 images respectively. We use the training set for pre-training the teachers. For TargetNet, we use the unlabeled images and the knowledge amalgamated from the teachers. The evaluation metric is Average Precision~(AP) and mean of AP (mAP) complying with the PASCAL challenge protocols.

As another popular benchmark for multi-label learning, We utilize Microsoft Common Objects in Context (MS-COCO)~\cite{Lin2014Microsoft} to validate the effectiveness of the proposed method, which contains 123,287 images for training and validation with 80 object concepts annotated. As the ground truth labels of the test set are not available, we evaluate the performance of all the methods on the validation set instead.

\textbf{Implementation Details.}
We implemented our model using TensorFlow with a NVIDIA M6000 of 24G memory.
We adopt the poly learning rate policy.
We set the base learning rate to 0.01, the power to 0.9,
and the weight decay to $5e-3$.
Due to limited physical memory on GPU cards and to assure effectiveness, we set the batch size to be 16 during training. The TargetNet which is in the same architecture as the teachers (ResNet50 and DenseNet), is initialized with parameters pre-trained on ImageNet.

\subsection{Experimental Results}

\textbf{Customized labels}. There are a total 20 labels in Pascal VOC 2007, we divide them randomly into two groups which are learned separately in two teacher networks. From each teacher, we choose one task to customize TargetNet dealing with two-label classification. The teachers are pre-trained in ResNet-50 with 10 labels. Here we conduct two sets of experiments: one is the two visual-irrelated tasks (`Bus' from teacher1, `Diningtable' from teacher2), the other one is the two visual-related tasks (`Dog' from teacher1, `Horse' from teacher2).

\begin{figure}[t]
\centering
\includegraphics[scale = 0.53]{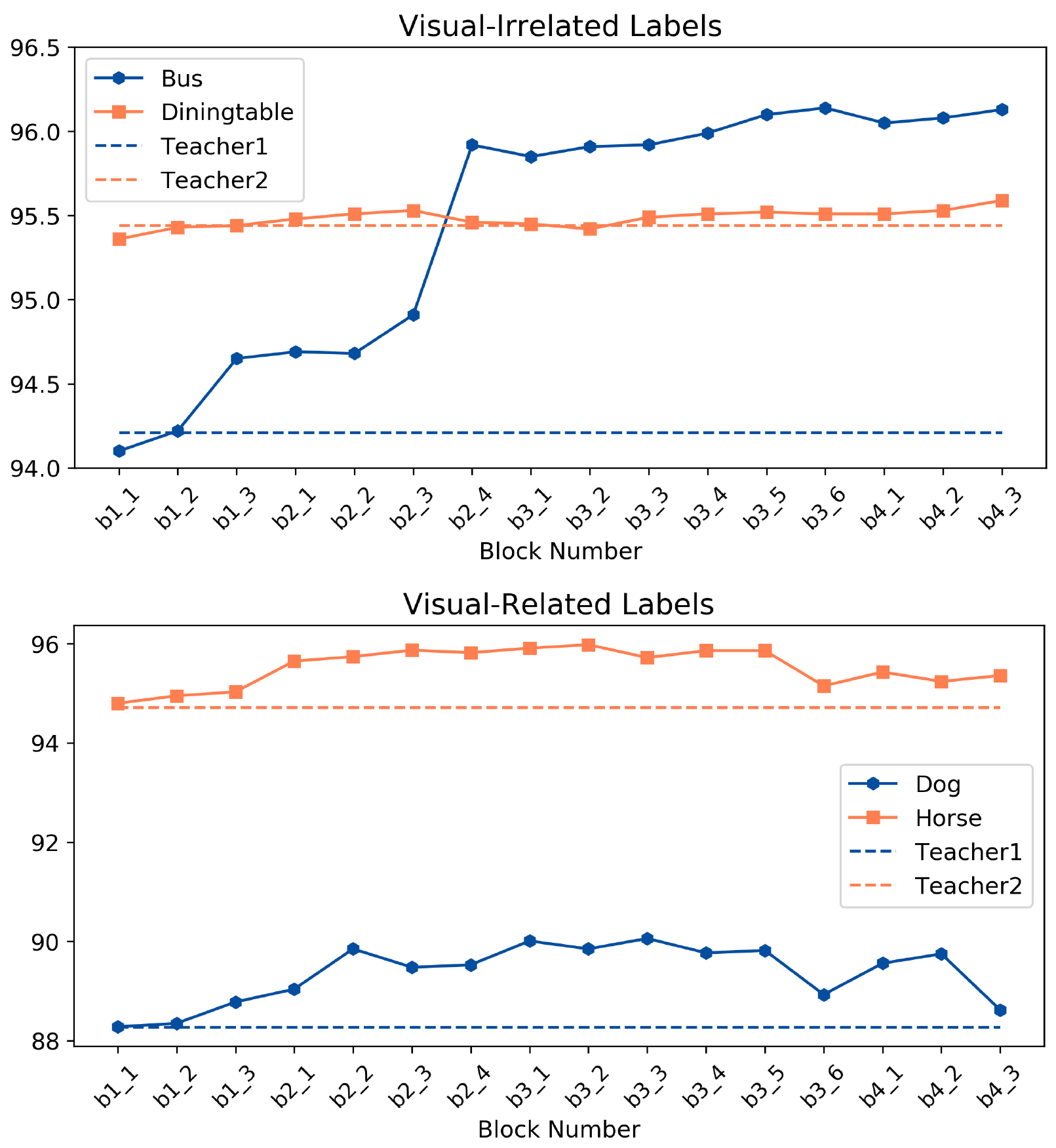}
\caption{The AP scores of TargetNet (ResNet-50) that branches out at different blocks.}
\vspace{2mm}
\label{fig:2-label}
\end{figure}

In Fig.~\ref{fig:2-label}, we show the accuracies of each task branching out at each block of ResNet-50. The upper one shows the trend of AP of two visual-irrelated labels, where we also show the initial AP of the label pre-trained in the teachers (`teacher1' and `teacher2'). For `bus', it achieves the highest AP of $96.2\%$ in block \textbf{b3\_6} surpassing the baseline in teacher1 with $94.2\%$, and TargetNet branches out at this block according to Eq.~(\ref{eq:branchout}), which conforms to the demand for branching out to get the highest accuracy for the chosen task. For `Dingingtable', TargetNet branches out at block \textbf{b4\_3} acquiring AP of $95.6\%$ in contrast with $95.4\%$ in teacher2.

In the lower chart of Fig.~\ref{fig:2-label}, the results of TargetNet customized with two visual-related labels are depicted. TargetNet branches out at \textbf{b3\_5} for `dog', \textbf{b3\_2} for `horse' with AP of $\{90.5\%,96.0\%\}$ vs $\{88.2\%,94.7\%\}$ in teachers. As can be seen in the results, classifications of `bus' and `Diningtable' share 13 blocks in TargetNet more than 9 blocks in classifications of `dog' and `horse', which shows the `visual-irrelated' set is more entwined than the `visual-related' one. This observation reveals that the relationship of multiple tasks is hard to determine and obtain, and the proposed method learns their hierarchy during the block-wise training.

\begin{table*}[t]
\centering
\caption{Comparisons of the classification results (AP in \%) with state-of-the-art approaches on VOC 2007}
\scriptsize
\begin{tabular}{p{15.5mm}p{3.2mm}p{3.2mm}p{3.2mm}p{3.2mm}p{3.2mm}p{3.2mm}p{3.2mm}p{3.2mm}p{3.2mm}p{3.2mm} p{3.2mm}p{3.2mm}p{3.2mm}p{3.2mm}p{3.2mm}p{3.2mm}p{3.2mm}p{3.2mm}p{3.2mm}p{3.2mm}p{3.2mm}}
\toprule
 Scenario  & plane & bike & bird& boat & bottle & bus & car & cat & chair& cow& table& dog& horse& motor &person &plant &sheep &sofa& train &tv &mAP\\
\midrule
TeacherRes50   &\bg 95.6 &\bg 86.0&\bg 95.0&\bg 96.3&94.3&\bg 94.7&\bg 92.9&91.8&89.3&96.5&95.0&88.2&\bg 96.7&\bg 94.9&\bg 68.1&94.0&97.4&93.9&\bg 95.5&94.6&92.5\\
TargetRes50\_b1&\bg 96.3&\bg 94.4&\bg 92.9&\bg 96.8&94.9&\bg 96.3&\bg 87.0&91.8&90.4&97.0&95.0&91.0&\bg 94.9&\bg 95.5&\bg 69.2&95.4&97.4&94.7&\bg 94.7&93.3&93.0 \\
TargetRes50\_b2&\bg 96.7&\bg 95.1&\bg 92.6&\bg 96.7&92.4&\bg 96.3&\bg 84.8&92.8&88.2&96.5&94.9&90.4&\bg 95.3&\bg 95.8&\bg 68.9&95.2&97.7&94.1&\bg 95.7&94.1&92.8\\
TargetRes50\_b3&\bg 97.0&\bg 82.3&\bg 93.9&\bg 97.1&95.2&\bg 96.4&\bg 83.4&89.7&89.6&97.2&95.5&91.5&\bg 92.4&\bg 94.9&\bg 68.7&94.5&98.1&94.7&\bg 96.0&94.1&92.1\\
TargetRes50\_b4&\bg 97.2&\bg 93.1&\bg 94.1&\bg 97.3&94.9&\bg 96.4&\bg 84.8&93.2&91.0&97.1&96.0&91.3&\bg 96.0&\bg 96.2&\bg 69.5&94.9&97.9&95.0&\bg 95.7&95.1&93.4\\
\midrule
TargetRes50&97.2&95.0&94.1&\textbf{97.3}&\textbf{95.2}&96.4&87.0&93.2&91.0&\textbf{97.1}&\textbf{96.0}&91.2&96.0&96.2&69.5&95.4&98.1&95.0&96.0&95.1&93.5\\
TargetRes50-3&96.5&95.4&93.9&95.5&94.0&\textbf{97.1}&90.4&92.3&\textbf{91.8}&\textbf{97.1}&95.3&87.8&95.6&96.0&75.9&95.3&97.2&95.5&96.8&94.9&\textbf{93.8}\\
TargetDenseNet&96.6&95.6&94.2&\textbf{97.3}&94.6&94.2&87.7&92.2&91.3&96.6&95.4&90.9&96.1&95.9&71.8&95.5&97.9&95.2&\textbf{95.5}&\textbf{95.3}&93.6\\
\midrule
\midrule
HCP-2000&96.0&92.1&93.7&93.4&58.7&84.0&93.4&92.0&62.8&89.1&76.3&91.4&95.0&87.8&93.1&69.9&90.3&68.0&96.8&80.6&85.2 \\
HCP-VGG&98.6&97.1&98.0&95.6&75.3&94.7&95.8&97.3&73.1&90.2&80.0&97.3&96.1&94.9&96.3&78.3&94.7&76.2&97.9&91.5&90.9\\
{RLSD+ft-RPN}&96.4&92.7&93.8&94.1&71.2&92.5&94.2&95.7&74.3&90.0&74.2&95.4&96.2&92.1&97.9&66.9&93.5&73.7&97.5&87.6&88.5 \\
CNN+LSTM&98.6&97.4&96.3&96.2&75.2&92.4&96.5&97.1&76.5&92.0&87.7&96.8&\textbf{97.5}&93.8&\textbf{98.5}&81.6&93.7&82.8&98.6&89.3&91.9 \\
PF-DLDL&\textbf{99.3}&\textbf{97.6}&\textbf{98.3}&97.0&79.0&95.7&\textbf{97.0}&\textbf{97.9}&81.8&93.3&88.2&\textbf{98.1}&96.9&\textbf{96.5}&98.4&84.8&94.9&82.7&98.5&92.8&93.4\\
\bottomrule
\end{tabular}
\label{tab:voc07}
\end{table*}

 \begin{table*}
 \centering
 \caption{Comparisons of the classification results (AP in \%) with state-of-the-art approaches on VOC 2012}
 \scriptsize
 \begin{tabular}{p{15.5mm}p{3.2mm} p{3.2mm} p{3.2mm} p{3.2mm} p{3.2mm} p{3.2mm} p{3.2mm} p{3.2mm} p{3.2mm} p{3.2mm} p{3.2mm} p{3.2mm} p{3.2mm} p{3.2mm} p{3.2mm} p{3.2mm} p{3.2mm} p{3.2mm} p{3.2mm}p{3.2mm}p{3.2mm}}
 \toprule
 Scenario  & plane & bike & bird& boat & bottle & bus & car & cat & chair& cow& table& dog& horse& motor &person &plant &sheep &sofa& train &tv &mAP\\
\midrule
HCP-2000&97.5&84.3&93.0&89.4&62.5&90.2&84.6&94.8&69.7&90.2&74.1&93.4&93.7&88.8&93.3&59.7&90.3&61.8&94.4&78.0&84.2 \\
Zhou et al.&96.3&84.2&90.3&57.3&86.1&90.9&86.2&94.3&67.2&82.0&91.4&94.4&88.1&94.5&64.9&82.4&76.8&74.1&96.6&82.0&84.0\\
HCP-VGG & 99.1&92.8&97.4&94.4&79.9&93.6&89.8&\textbf{98.2}&78.2&94.9&79.8&97.8&97.0&93.8&96.4&74.3&94.7&71.9&96.7&88.6&90.5\\
PF-DLDL&\textbf{99.5}&94.1&\textbf{97.9}&95.9&81.0&94.8&\textbf{93.1}&\textbf{98.2}&82.4&96.1&84.0&\textbf{98.0}&\textbf{97.8}&95.7&\textbf{97.7}&78.9&95.5&78.0&97.8&92.2&92.4\\
 \midrule
TargetRes50&95.8&95.0&91.0&95.0&92.4&\textbf{97.1}&89.3&88.5&88.4&96.8&95.1&86.0&94.7&94.9&70.8&94.7&96.8&94.9&96.1&93.9&92.5\\
TargetRes50-3&96.4&\textbf{95.4}&92.9&95.3&\textbf{93.9}&97.0&90.2&89.3&\textbf{89.8}&\textbf{97.0}&95.1&87.7&95.6&\textbf{95.9}&72.7&\textbf{95.3}&96.9&\textbf{95.4}&96.6&\textbf{94.8}&93.2\\
TargetDenseNet&96.2&95.3&94.9&\textbf{96.8}&92.1&96.4&91.0&92.1&85.7&\textbf{97.0}&\textbf{95.2}&88.9&96.2&95.8&80.5&93.3&\textbf{97.0}&90.2&\textbf{97.9}&94.6&\textbf{93.3} \\
 \bottomrule
 \end{tabular}
 \label{tab:voc12}
 \end{table*}

\textbf{Multiple Labels}.
To the best of our knowledge, we are the first to study the customized-label classification. So we set the customized task set $\mathcal{C}=\mathcal{A}$, which makes the TargetNet deal with a normal multi-label task for comparison with other methods. We compare the performance of the proposed method against the following state-of-the-art approaches: HCP-2000~\cite{wei2014CNN}, HCP-VGG~\cite{wei2016hcp:}, RLSD+ft-RPN~\cite{zhang2018multilabel}, CNN+RNN~\cite{wang2016cnn}, CNN+LSTM~\cite{wang2017multi-label}, PF-DLDL~\cite{gao2017deep} and the method of~\cite{zhou2018an}.

Tab.~\ref{tab:voc07} displays the results on Pascal VOC 2007 dataset. `TargetRes50' indicates the TargetNet in ResNet-50 and branches out at different blocks for each task to obtain the best accuracy, outperforming the teachers `TeacherRes50', which consists of two teachers each specifying in a 10-label classification task, which are distinguished by different background colors in the tale. Also we show the APs in TargetNet of 4 main blocks during the block-wise training as `TargetRes50\_b1$\sim$b\_4', where the most of the labels' AP scores are increased. Also, we divide the labels into 3 groups, each learned in one teacher network. Then `TargetRes50-3' is the TargetNet guided by these 3 teachers, which increases the accuracies further than that guided by two teachers. For the generalization test, we change the architecture to DenseNet `TargetDenseNet' for both the target and teachers, and acquires better results thanks to better teachers' performance.

Tab.~\ref{tab:voc12} reports details of all experimental results on Pascal VOC 2012 dataset. Our work, TargetNets amalgamated from 2 teachers (`TargetRes50' and `TargetDenseNet') and amalgamated from 3 teachers (`TargetRes50-3') all outperform the other methods with the highest mAP of $93.3\%$ in `TargetDenseNet',  note that we even do not use the image region information.

Tab.~\ref{tab:coco} presents the results on MS-COCO dataset. We follow the work of ~\cite{wang2016cnn} to select the top $k = 3$ labels for each image. In tab.~\ref{tab:coco}, we compare the overall precision (`O-P'), recall (`O-R'), F1 (`O-F1'), and per-class precision (`C-P'), recall (`C-R'), F1 (`C-F1'). Our proposed TargetNets learned from two teachers (`TargetRes50' and `TargetDenseNet') outperform the exiting methods and achieve the best on all the measurements. And `TargetDenseNet' works better than `TargetRes50' in 5 indicators.

 \begin{table}
 \centering
 \scriptsize
 \caption{Comparison of our method and state-of-the-art methods on the MS-COCO dataset}
\begin{tabular}{lcccccc}
 \toprule
Method  & C-P & C-R & C-F1 & O-P&O-R&O-F1 \\
 \midrule
RLSD+ft-RPN &67.6&57.2&62.0&70.1&63.4&66.5\\
CNN+RNN & 66.0&55.6&60.4&69.2&66.4&67.8     \\
CNN+LSTM & 79.1&58.7&67.4&84.0&63.0&72.0      \\
\midrule
TargetRes50&79.8&63.5& \textbf{69.4}& 83.9& 65.7&75.0\\
TargetDenseNet & \textbf{80.5}&\textbf{64.0}&69.3&\textbf{85.4}&\textbf{66.9}&\textbf{75.2} \\
\bottomrule
\end{tabular}
\label{tab:coco}
\end{table}

\section{Conclusion and Future Work}

In this paper, we focus on training a customized multi-task student model, by learning from a pool of multi- or single-task teacher models and without any human-labelled annotation.
The student amalgamates the relevant knowledge filtered from the teachers,
and masters the complete or a subset of expertise of them.
Specifically, we adopt a block-wise scheme that trains the features of the student entangled with the teachers.
Our experimental results demonstrate that, the derived student model achieves very promising results without any groundtruth annotations, even outperforms those of the teachers in their own domain.

The current work, admittedly, focuses on the classification task only. In the future, our goal is to
conduct knowledge amalgamation from
multiple heterogeneous networks handling other tasks,
and make it generalized to an even broader domain. This includes
not only the higher-level vision tasks including object detection~\cite{wang2018cvpr}, tracking~\cite{wang2014eccv}, and image search~\cite{wang2011tip},
but also the lower-level ones like image rectification~\cite{wang2018eccv}.

\section*{Acknowledgement}
This work is supported by  National Key Research and Development Program (2016YFB1200203) , National Natural Science Foundation of China (61572428,U1509206), Key Research and Development Program of Zhejiang Province (2018C01004), the Program of International Science and Technology Cooperation (2013DFG12840), the Startup Funding of Stevens Institute of Technology and the Fundamental Research Funds for the Central Universities.


\end{document}